\documentclass{ifacconf}

\usepackage{graphicx}      
\usepackage{natbib}        
\usepackage{amsmath,amssymb}
\usepackage{wrapfig}
\usepackage{float}
\usepackage{tikz}
\usepackage{pgfplots}
\usepackage{svg}
\usepackage{ragged2e}
\usepackage[utf8]{inputenc}
\DeclareUnicodeCharacter{2212}{−}
\usepgfplotslibrary{groupplots,dateplot,fillbetween}
\usetikzlibrary{patterns,shapes.arrows}
\pgfplotsset{compat=newest}
\usepackage{nicefrac}
\usepackage{comment}
\usepackage{color}

\begin{document}
\begin{frontmatter}

\title{Customizing Textile and Tactile Skins for Interactive Industrial Robots}


\author[First]{Bo Ying Su}
\author[First]{Zhongqi Wei} 
\author[First]{James McCann}
\author[First]{Wenzhen Yuan}
\author[First]{Changliu Liu}

\address[First]{Carnegie Mellon University (emails: boyings, zhongqi2, jmccann, wenzheny, cliu6@andrew.cmu.edu)}

\begin{abstract}                
Tactile skins made from textiles enhance robot-human interaction by localizing contact points and measuring contact forces. This paper presents a solution for rapidly fabricating, calibrating, and deploying these skins on industrial robot arms. The novel automated skin calibration procedure maps skin locations to robot geometry and calibrates contact force. Through experiments on a FANUC LR Mate 200id/7L industrial robot, we demonstrate that tactile skins made from textiles can be effectively used for human-robot interaction in industrial environments, and can provide unique opportunities in robot control and learning, making them a promising technology for enhancing robot perception and interaction.
\end{abstract}


\end{frontmatter}
\begin{figure*}[t]
  \begin{center}
  \centering
  \includegraphics[width=7in]{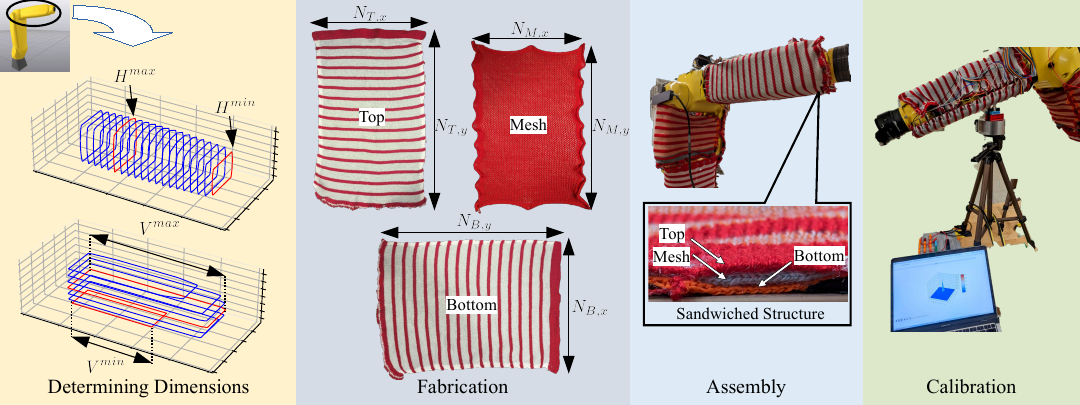}
  \caption{The fabrication process for the textile and tactile skin. Firstly, the dimensions of the top, mesh, and bottom layers of the skin are determined by measuring $H^{min}$, $H^{max}$, $V^{min}$, and $V^{max}$ from each link of the robot mesh. 
  Next, the three layers of the skin are knitted and assembled by stacking them together to form a pressure-sensitive array. Finally, the skin is attached to the robot, and a calibration process is performed to determine the position and force responses of individual sensing cells.}
  \label{fig: fabrication}
  \end{center}
\end{figure*}

\section{Introduction}


Industrial robots usually perform precise and repetitive tasks, but are limited by highly structured and deterministic environments, hindering their flexibility. In order for robots to adapt to changing environments and perform unstructured tasks, it is critical to improve their interactivity. This paper introduces textile and tactile skins as a new sensing modality, and demonstrates their effectiveness in enhancing interactivity of industrial robots.

Robot control in interactive tasks heavily relies on available sensors. Six-axis Force Torque Sensors (FTS) can measure contact force and torque but cannot localize the contact position~\citep{cao2021six}. Robot joint motor measurements can estimate external force and detect collisions, but do not localize the contact positions~\citep{wahrburg2017motor}. Vision-based systems can localize contact but do not detect force and suffer from occlusion~\citep{kloss2020accurate}. Light curtains do not suffer from occlusion but can not be easily retrofitted onto complex geometries~\citep{tsuji2020general}. On the other hand, FTS can provide real-time data with high frequency, while vision-based systems are limited by the camera frame rate, typically 30 Hz or lower. FTS require professional installation, while vision-based systems are more portable but require significant calibration effort.


Tactile skins are a type of sensors that can be utilized in robots for a variety of control applications~\citep{dahiya2013directions}. Previous researchers have successfully developed tactile skins for human-robot interactive tasks~\citep{cirillo2015conformable, baykas2020safe, fan2022enabling}. However, these sensors are often not customizable and are difficult to adapt to different robots. Additionally, while some commercial skins, such as the T-skin~\citep{examplewebsite}, are available to enhance robot safety, they are in general costly. Moreover, relying solely on basic safety settings where robots stop upon skin contact may not be suitable for intricate human-robot interaction tasks.

Recently, a new type of flexible tactile skin using textiles called the Robot Sweater has been proposed~\citep{si2023robotsweater}. These soft and deformable sensors can be easily deployed to various robots by creating customized knitting patterns. The tactile skin detects contact forces and can localize multiple contact positions. As it is attached directly to the robot surface, it is free from occlusion issues. The tactile skin provides real-time data with a high frequency of up to 150 Hz and is deformable and stretchable, making it easy to be deployed to different robots by adjusting its dimensions and tension to achieve a tight and robust fit to the robot surface, even with complex curvature.


However, these tactile skins have yet to be applied to industrial robots, which require higher precision in localization and pressure sensing for control applications. To address this gap, this paper presents a holistic solution for rapidly fabricating, calibrating, and deploying tactile skins to industrial robot arms. Our contributions are as follows: 1) we introduce a method for determining skin parameters based on robot geometry (Section 2); 2) we propose an automated skin calibration process that calibrates both the position and force reading from the skin to the robot (Section 3); and 3) we demonstrate how the skins can be integrated into robot control and learning to improve interactivity of industrial robots through experiments on a FANUC LR Mate 200id/7L industrial robot arm (Section~4).


\section{Skin Fabrication}
This section first reviews the principles and sensing mechanisms of the textile and tactile skin. We then discuss how to determine the appropriate skin parameters for different robots.

\subsection{Principles of the Textile and Tactile Skin}
As proposed by~\cite{si2023robotsweater}, the textile and tactile skin, Robot Sweater, consists of three layers of fabric in a sandwich structure, as shown in 
 the assembly stage of Figure \ref{fig: fabrication}. The top and bottom layers contain alternating conductive and non-conductive stripes placed orthogonally, while the middle layer is formed by a mesh that separates the top and bottom layers. When pressure is applied to the skin, the top and bottom layers are stretched, causing the conductive stripes to touch each other and complete the circuit. The resistance, which is inversely correlated to the touching area of the top and bottom layers, is then recorded by an Arduino microcontroller. Once the skin parameters are determined, the skin can be automatically made by a knitting machine. 

\subsection{Determination of Skin Parameters}
To successfully use tactile skins on industrial robot arms, we must choose the appropriate skin material and dimensions. We will explain how to determine these parameters for a given robot.

\begin{figure}[t]
  \begin{center}
  \centering
  \includegraphics[width=0.9\linewidth]{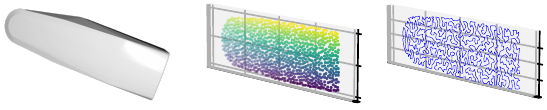}
  \caption{Planning the skin calibration sequence. Firstly, the surface of the robot link is sampled from the mesh. Next, a trajectory is planned for the force-torque sensor to visit each sample on the surface. The robot executes the trajectory while the force-torque sensor records the force and torque at each sample point. The robot touches each location of the skin with varying pressure from 0N to 10N.}
  \label{fig: calibration}
  \end{center}
\end{figure}

\begin{figure}[!t]
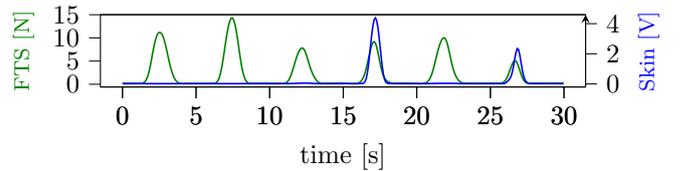

  \centering
\include{calib_sample_data}%
\vspace{-30pt}
\caption{Sample data collected during the calibration process for cell 80 of the tactile skin. The calibration process records force torque sensor readings, the robot's joint angle, and the skin's pressure readings simultaneously. This figure shows that certain points on the robot surface trigger a response from cell 80, while others do not. These responses allow us to localize the position of cell 80 on the robot surface.}
  \label{fig: sample_data}
\end{figure}

\subsubsection{Skin Material}
In \cite{si2023robotsweater}, Rayon yarn was used in the skin's fabrication. In this work, we found that acrylic yarn is more robust when attached to complex robot geometries due to its thickness and it being sturdier.

\subsubsection{Skin Dimensions}
This work uses a flat rectangular skin to wrap around a cylinder-link object for ease of fabrication. 
For a cylinder-like object (shown in the first stage of Figure \ref{fig: fabrication}), we extract the following parameters: 1) the longest and shortest perimeters along the rotational x-axis: $H^{max}$ and $H^{min}$; 2) the maximum and minimum partial height along the orthogonal y-axis: $V^{max}$ and $V^{min}$.

The parameters we need to choose are the number of columns (x) and rows (y) of the top (T), mesh (M), and bottom (B) layers of the skin, denoted by $N_{i,j}\in\mathbb{N}$, where $i \in{T,M,B}$ and $j\in{x,y}$. Note that the actual size of the skin layer $i$ is determined by the stitch conversion ratio: $R_{i,x}$ (horizontal ratio) and $R_{i,y}$ (vertical ratio) for $i\in{T,M,B}$.\footnote{The stitch conversion ratio is 0.889 stitches/mm for the x-axis and 0.981 stitches/mm for the y-axis for the top and bottom layers, while it is 0.543 stitches/mm for the x-axis and 0.437 stitches/mm for the y-axis for the mesh layer.} The non-stretched dimension of the skin layer $i$ is $\nicefrac{N_{i,x}}{R_{i,x}}\times \nicefrac{N_{i,y}}{R_{i,y}}$.

Since the knitted skin has different stretchability along the horizontal and vertical axes, we define the maximum horizontal stretch as $\bar{S}_{i,x}$ and the maximum vertical stretch as $\bar{S}_{i,y}$ for $i\in\{T,M,B\}$. Meanwhile, to obtain reasonable conductivity on the conductive stripes as well as to tightly fit the cylinder-like object, the skin needs to be stretched to some extent. We define the minimum horizontal stretch as $\underline{S}_{i,x}$ and the minimum vertical stretch as $\underline{S}_{i,y}$ for $i\in\{T,M,B\}$.\footnote{According to our tuned stitch sizes, the top and bottom layer skin can stretch up to 177\% of their original length horizontally and up to 145\% vertically. 
To obtain reasonable conductivity on the conductive stripes, the top and bottom layer skin must be stretched to at least 145\% of their original length horizontally. There is no such limitation on the vertical axis, though. The implication here is that for the top and bottom layers, the range of 145\% to 177\% of the original length should contain $H^{max}$ and $H^{min}$; and the range between 100\% to 145\% should contain $V^{max}$ and $V^{min}$. For the mesh layer, there is no such limitations.} 
Hence, the actual size of the skin layer $i$ should be within $[\nicefrac{N_{i,x}\underline{S}_{i,x}}{R_{i,x}}, \nicefrac{N_{i,x}\bar{S}_{i,x}}{R_{i,x}}]\times [\nicefrac{N_{i,y}\underline{S}_{i,y}}{R_{i,y}}, \nicefrac{N_{i,y}\bar{S}_{i,y}}{R_{i,y}}]$. To let the skin fit the cylinder-like object, the following relationship needs to be satisfied $\forall i\in\{T,M,B\}$,
\begin{subequations}\label{eq: dimension condition}
\begin{align}
\nicefrac{N_{i,x}\underline{S}_{i,x}}{R_{i,x}}\leq H^{min}\leq H^{max} \leq \nicefrac{N_{i,x}\bar{S}_{i,x}}{R_{i,x}}, \\
\nicefrac{N_{i,y}\underline{S}_{i,y}}{R_{i,y}} \leq V^{min}\leq V^{max} \leq \nicefrac{N_{i,y}\bar{S}_{i,y}}{R_{i,y}}
\end{align}
\end{subequations}

In order to ensure feasibility of \eqref{eq: dimension condition}, the object must satisfy that
$\nicefrac{H^{max}}{H^{min}} \leq \min_{i\in\{T,M,B\}}\{\nicefrac{\bar{S}_{i,x}}{\underline{S}_{i,x}}\}$ and $
  \nicefrac{V^{max}}{V^{min}} \leq \min_{i\in\{T,M,B\}}\{\nicefrac{\bar{S}_{i,y}}{\underline{S}_{i,y}}\}$, 
so that the maximum perimeter of the object falls within the skin's stretch tolerance. 

Then we select the smallest dimension that satisfies \eqref{eq: dimension condition}, where for $i\in\{T,M,B\}$,
\begin{subequations}
\begin{align}
    N_{i,x} = \lceil \nicefrac{H^{min}R_{i,x}}{\underline{S}_{i,x}}   \rceil\\
    N_{i,y} = \lceil \nicefrac{V^{min}R_{i,y}}{\underline{S}_{i,y}}   \rceil.
\end{align}
\end{subequations}

\begin{figure}[t]
  \begin{center}
  \centering
  \includegraphics[width=\columnwidth]{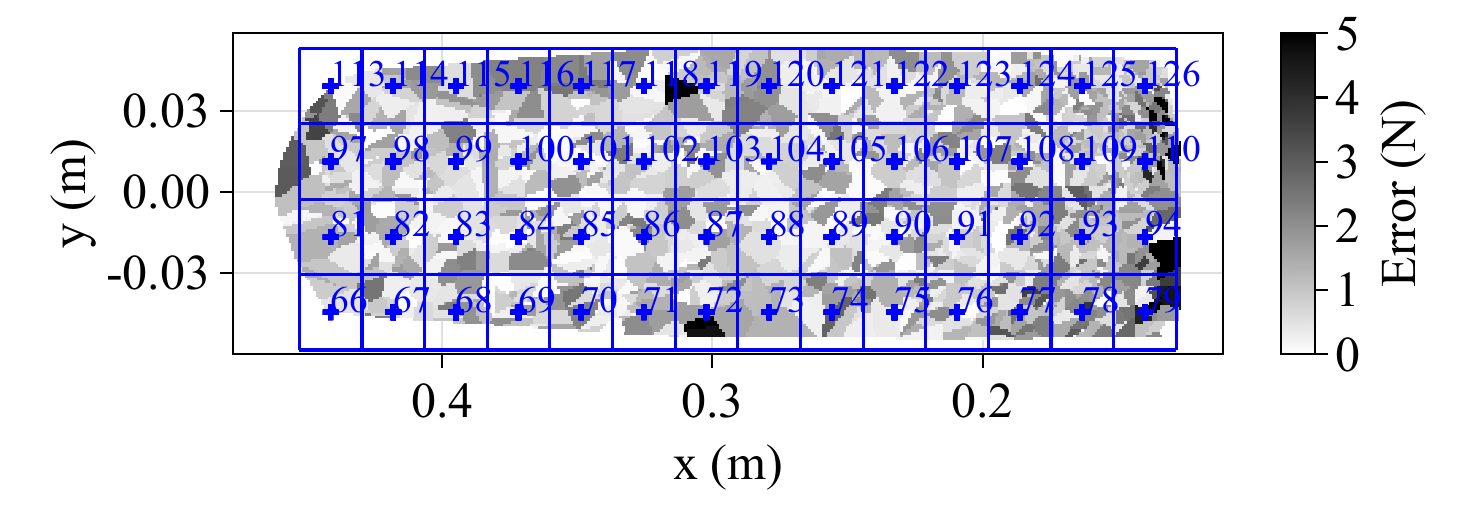}
  \includegraphics[width=\columnwidth]{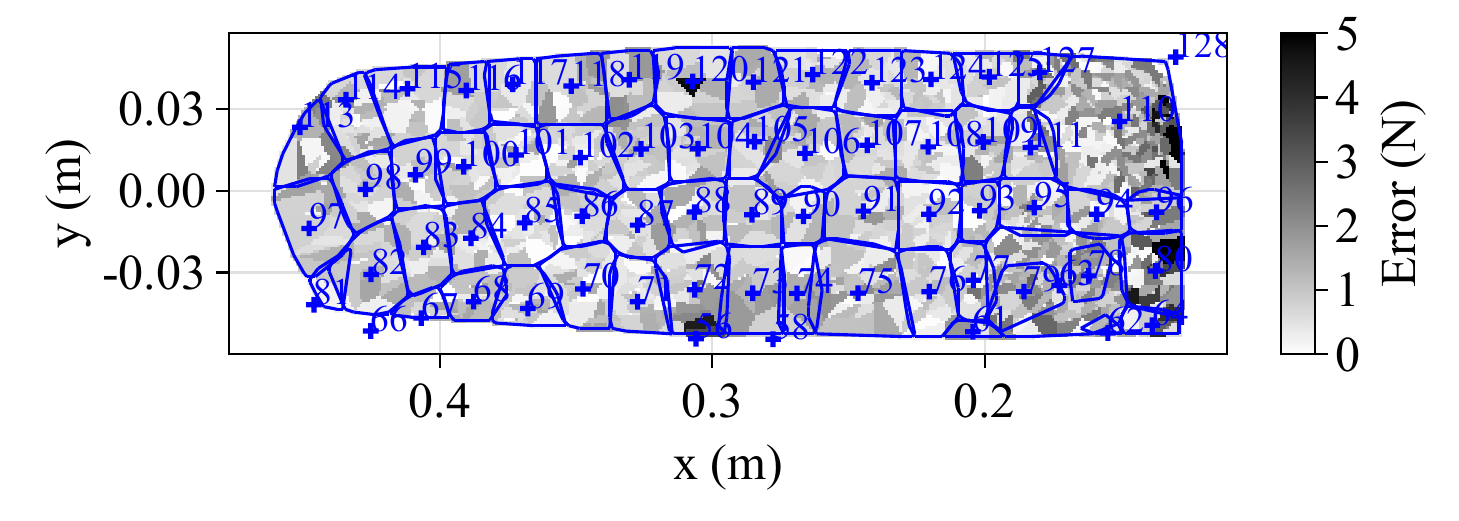}
\caption{Calibration result. The first plot shows the localization and force estimation error before calibration. The second plot shows the calibration result. The blue boundaries show the calculaed receptive field of the sensing cells. The darker color represents a greater force estimation error compared to the ground truth.}
  \label{fig: calibration}
  \end{center}
\end{figure}

\section{Modeling and Calibration}

This section discusses how to calibrate both the location and the pressure response of the skin after installation. As shown in the calibration stage of Figure \ref{fig: fabrication}, the skin is stretched and can be slightly distorted when attached to the robot, making position calibration necessary to accurately determine the receptive field and force response of sensing cells. We accomplish this by repeatedly touching a fixed force torque sensor with the skin, as shown in the calibration stage of Figure \ref{fig: fabrication}. The entire process can be done automatically.

\subsection{Data Collection for Calibration}

To localize the skin and calibrate pressure measurement, we use an ATI Omega 85 force torque sensor attached to a fixed tripod (see calibration stage in Figure \ref{fig: fabrication}). Our aim is to obtain ground truth force torque signals for nearly every point on the skin. Data collection involves the robot approaching the sensor and actively touching it with the skin, generating a skin-traversal trajectory through Poisson-disc sampling on the target robot link's surface. The sample distance is half the grid size to ensure calibration of all cells during the run. The grid size is determined by the skin dimensions and the number of available analog Arduino pins (16 in our case using Arduino Mega 2560 Rev3). We allow for a maximum 26-degree angle between the force torque sensor and skin surface normal; and it is compensated through force vector projections. The order to traverse all points is optimized using a traveling salesman solver (\cite{ortools}) to minimize the calibration trajectory's make span. We collect timestamped tuples $(\forall i, C_{i,t}, F_t, P_t)$ at approximately 43 frames per second, where $i\in\mathbb{N}$ denotes the sensing cell, $C_{i,t}\in\mathbb{R}$ denotes the skin response at cell $i$ at time $t$, $F_t\in\mathbb{R}$ represents the force torque sensor reading at time $t$, and $P_t\in\mathbb{R}^3$ indicates the force torque sensor position relative to the link origin at time $t$. Figure \ref{fig: sample_data} illustrates an example data sequence.

\subsection{Skin Localization}
To determine the location of a sensing cell relative to the robot, we use a one-to-rest k-nearest neighbor classifier on the collected data.  The k-nearest neighbor classifier has been shown to have robust performance in localizing the contact position of tactile sensors \citep{moham2019sensors}. We apply a similar technique, but instead of determining the actual contact location, we calculate the receptive field of individual sensing cells based on the ground truth contact location given by the robot's forward kinematics.

The k-nearest neighbor algorithm calculates the weighted distance between the force torque sensor position $P_t$ and the skin response $C_{i,t}$ of each sensing cell. We use weighted distance, where the weight assigned to each neighbor is inversely proportional to its distance from the query point, denoted as $d_i$. This is because our sampling points are sparse across the surface, and using weighted distance reduces the influence of noisy neighbors in the prediction. The prediction is then given as the class with the most votes from its k nearest neighbors.
\begin{equation}
  \hat{y} = \arg\max_{y} \sum_{i=1}^k w_i \delta(y, y_i)
\end{equation}
where $w_i = 1 /d_i^2$ is the weight of the $i^{th}$ nearest neighbor, and $\delta(y, y_i)$ is the indicator function that is 1 if $y = y_i$ and 0 otherwise. 

We design k nearest neighbor classifiers for each cell. We selected k = 1000 as a hyperparameter since it provides the most robust outlier rejection. Each of the classifiers predicts whether a contact at a position in space could be detected by the cell. We then created an ensemble classifier which determines the receptive fields of each cell using majority voting. The classified cell boundaries and center locations are shown in Figure \ref{fig: calibration}.

\subsection{Pressure Calibration}

After skin localization, the force response of each cell can be calibrated using linear regression of the data $\{(F_t, C_{i,t})\}$. 
We can find a linear function $f_i: C_{i,t}\mapsto F_t$ that best fits the data points for each sensing cell $i$ by solving the following least square problem:
\begin{equation}
\min_{f_i} \sum_{t \text{, s.t.} i_t= i} (f_i(C_{i,t}) - y_t)^2.
\end{equation}

Figure \ref{fig: calibration} presents two plots representing the calibration  outcomes for contact and force. We assessed our calibration model's performance by splitting the dataset into a 90\% training set and a 10\% testing set, with the plots depicting the test error. The first plot showcases contact localization results, using the receptive field centroids of each cell to predict contact location based on cell output. Our testing set yielded a contact localization root mean square error (RMSE) of 3.00 cm. Before calibration, while assuming the skin is a perfect grid, the RMSE is 5.83 cm. Note that the localization error is limited by the grid sizes of the cells, which the mean side lengths of all cells is 2.89 cm. The second plot illustrates force calibration results, employing linear models for each cell to predict force based on cell voltage. The testing set exhibited a force prediction RMSE of 1.36 N, which is an improvement over naive scaling of skin readings based on cell's saturation force where the RMSE is 1.92 N.

\section{Control Applications}

\begin{figure}[t]
  \begin{center}
  \centering
  \includegraphics[width=1\linewidth]{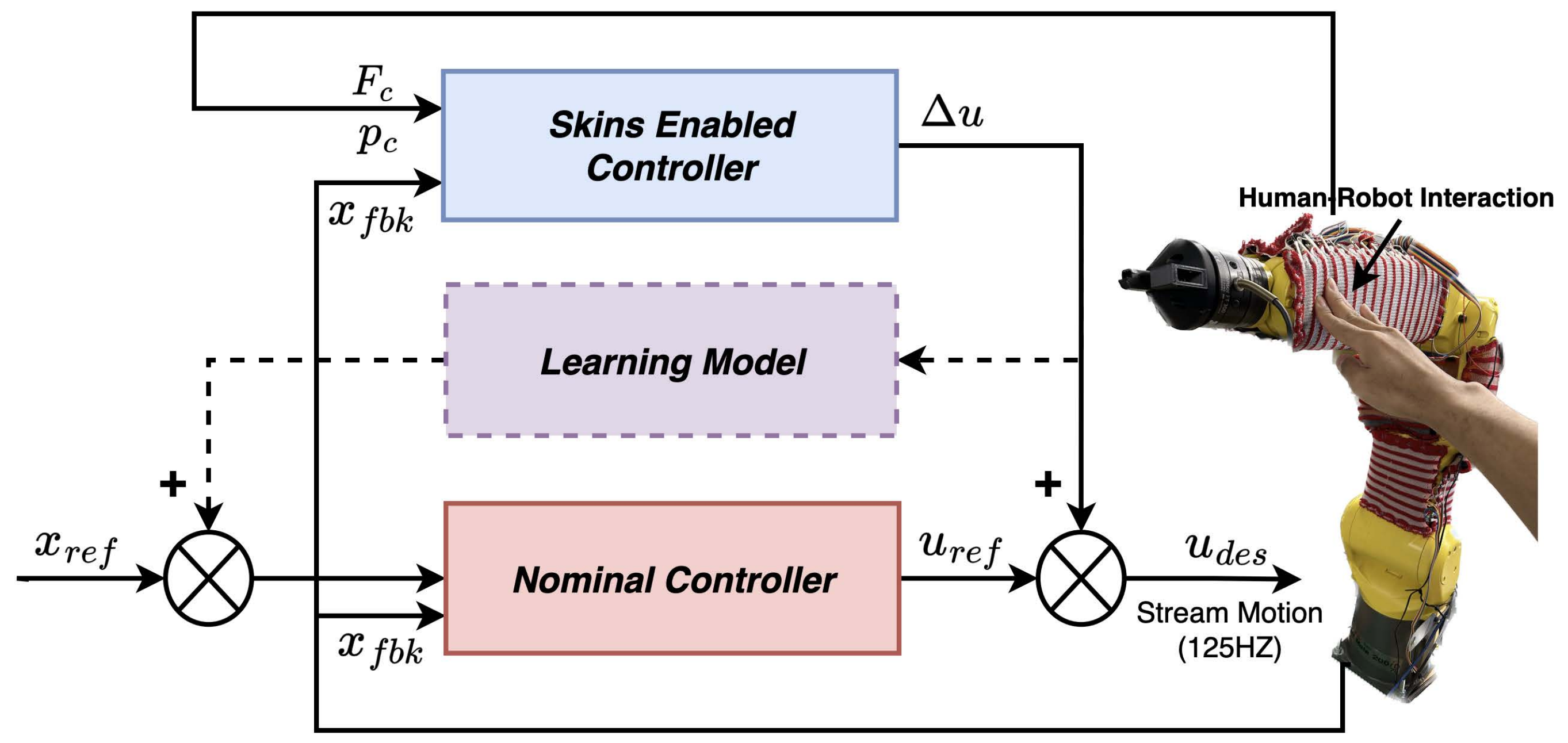}
  \caption{The proposed skin-based control and learning framework.}
  \label{fig: overview of learning from human contact}
  \end{center}
\end{figure}

This section discusses how to use the skin to facilitate robot learning from human contacts, as shown in Figure \ref{fig: overview of learning from human contact}. The skin-enabled controller uses contact force and position readings to calculate control displacements, which directly adjust robot behavior and provide rich signals for robot learning tasks. This process could allow for interactive responses to human contact and improve the overall effectiveness of human-robot interaction. The learning model is sensor-agnostic and has been studied by \cite{shek2022learning}. This section mainly focuses on the skin-enabled controller, with the integration of the learning model left for future work.

We provide two example design of the skin enabled controllers: 1) one controller that modifies robot trajectories using human contact signals and 2) one controller that regulates force interactions between human and robot using tactile skins. 

The experiments were conducted using the FANUC LR Mate 200iD/7L robot arm. We fabricated two rectangular skins for the second and fourth links, whose top layers measured 330 mm (width) by 363 mm (height) and 455 mm (width) by 140 mm (height), respectively. Each skin had an array of sensing cells, with the first skin containing a 16 $\times$ 16 array and the second skin containing a 16 $\times$ 8 array. The signals from each sensing cell were routed to an Arduino board, which then transmitted the signals to the controller via Ethernet. The transmission rate was 43 frames per second for the 16 $\times$ 16 skin and 74 frames per second for the 16 $\times$ 8 skin. The bandwidth of skin signals is limited by the hardware and will be addressed in future work. The control command was computed and sent to the robot via the stream motion interface at a rate of 125 Hz.

\subsection{Trajectory Modification from Human Feedback}

In this use case, we consider the situation when the human pushes the robot away from potentially dangerous configurations while the robot is following a Cartesian trajectory. 
Skin-enabled controllers can efficiently detect the contact position and generate new trajectories that minimize disruption to the robot's original motion. Inspired by the collision avoidance algorithm in~\cite{liu2017ccta}, we formulate the trajectory modification problem as a quadratic program (QP). The objective of this QP formulation is to minimize the velocity difference between the original and modified trajectories while ensuring that the robot follows human's commands. At the current joint configuration $q\in\mathbb{R}^6$, the velocity command $\dot q_{cmd}\in\mathbb{R}^6$ is computed by: 
\begin{equation}\label{eq: ccta}
  \begin{aligned}
    & \min_{\dot{q}_{\text{cmd}}}
    & & ||\dot{x}_{ref} - J(q) \dot{q}_{\text{cmd}}||^2_2 \\ 
    & \text{s.t.}
    & & {u_{\text{i}}}^\top J_{\text{i}} \dot{q}_{\text{cmd}} \geq f(F_{i}),\forall i
  \end{aligned}  
\end{equation}
where $J\in\mathbb{R}^{3\times 6}$ is the Jacobian matrix associated with the end effector $e$ at robot configuration $q$, $\dot{x}_{ref}\in\mathbb{R}^{3\times 1}$ is the reference velocity of the end effector, $u_{\text{i}}\in\mathbb{R}^{3\times 1}$ is the unit vector pointing away from the contact point $i$, and $J_{\text{i}}\in\mathbb{R}^{3\times 6}$ is the Jacobian matrix evaluated at the contact point $i$, $f(F_{i})$ is a non-negative and monotonic function that depends on the contact force at contact point $i$. When $f(F_{i})\equiv 0$, the constraint is simply requiring that there is no motion against human touch. When $f(F_{i})\propto F_{i}$, the robot will be pushed further away if the human pushes harder. 
The dimension of the constraint is the same as the number of contacts. Note that this formulation is different from \cite{liu2017ccta} in the following aspects: 1) \eqref{eq: ccta} allows contact while \cite{liu2017ccta} avoids contact; 2) the magnitude of the constraint in \eqref{eq: ccta} can be dependent on the magnitude of the contact force, while \cite{liu2017ccta} considers constant constraints in all situations, i.e., $f(F_{i})\equiv 0$.

\begin{figure}[t]
  \begin{center}
  \centering
  \begin{center}
  \centering
  \includegraphics[width=0.9\linewidth]{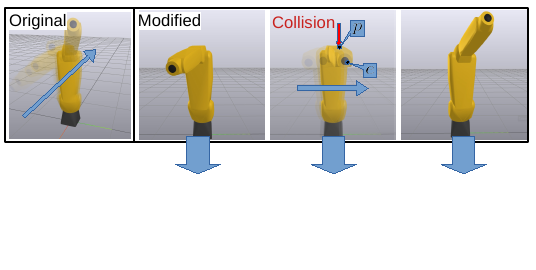}
\vspace{-55pt}
\begin{flushleft}
\begin{tikzpicture}

    \definecolor{darkgray176}{RGB}{176,176,176}
    \definecolor{steelblue31119180}{RGB}{31,119,180}
    
    \begin{axis}[
    tick align=outside,
    tick pos=left,
    scaled y ticks=false,
    height=2.5cm,
    yticklabel style={text width=3em, align=right, /pgf/number format/fixed},
    width=0.9\linewidth,
    x grid style={darkgray176},
    xmin=10, xmax=20,
    xtick=\empty,
    y grid style={darkgray176},
    ylabel={$C_{148}$ [N]},
    ytick={0,20},
    ymin=0, ymax=20,
    ]
\addplot [draw=none, fill=gray!50,opacity=0.5] coordinates {(13.8,-1) (13.8,25) (17,25) (17,-1)};
\addplot [semithick, steelblue31119180]
table {%
10.2398681640625 0
10.3598666191101 0
10.4798650741577 0
10.5998635292053 0
10.7198619842529 0
10.8398604393005 0
10.9598588943481 0
11.0798573493958 0
11.1998558044434 0
11.319854259491 0
11.4398527145386 0
11.5598511695862 0
11.6798496246338 0
11.7998480796814 0
11.919846534729 0
12.0398449897766 0
12.1598434448242 0
12.2798418998718 0
12.3998403549194 0
12.519838809967 0
12.6398372650146 0
12.7598357200623 0
12.8798341751099 0
12.9998326301575 0
13.1198310852051 0
13.2398295402527 0
13.3598279953003 0
13.4798264503479 0
13.5998249053955 0
13.7198233604431 0
13.8398218154907 0.0381119576816796
13.9598202705383 1.42002554062076
14.0798187255859 2.54148076859922
14.1998171806335 4.09922878463576
14.3198156356812 8.3345344982612
14.4398140907288 10.8119626344792
14.5598125457764 11.7864629893106
14.679811000824 7.55879928908515
14.7998094558716 3.15856070199935
14.9198079109192 4.91211147935803
15.0398063659668 10.3495670881951
15.1598048210144 15.4190463829162
15.279803276062 18.5521990126936
15.3998017311096 19.0786979450791
15.5198001861572 19.9444559643261
15.6397986412048 17.1371991144526
15.7597970962524 14.6967272965595
15.8797955513 13.3450112507446
15.9997940063477 13.1537886113456
16.1197924613953 14.599275283888
16.2397909164429 15.2529725105025
16.3597893714905 15.5482996081293
16.4797878265381 17.1018057950222
16.5997862815857 11.958893308897
16.7197847366333 7.09409384152623
16.8397831916809 1.1258750789806
16.9597816467285 0
17.0797801017761 0
17.1997785568237 0
17.3197770118713 0
17.4397754669189 0
17.5597739219666 0
17.6797723770142 0
17.7997708320618 0
17.9197692871094 0
18.039767742157 0
18.1597661972046 0
18.2797646522522 0
18.3997631072998 0
18.5197615623474 0
18.639760017395 0
18.7597584724426 0
18.8797569274902 0
18.9997553825378 0
19.1197538375854 0
19.2397522926331 0
19.3597507476807 0
19.4797492027283 0
19.5997476577759 0.00382176269150792
19.7197461128235 0.00382176269150792
19.8397445678711 0.00382176269150792
19.9597430229187 0
20.0797414779663 0
20.1997399330139 0
20.3197383880615 0
20.4397368431091 0
20.5597352981567 0
20.6797337532043 0
20.799732208252 0
20.9197306632996 0
};
\end{axis}
\end{tikzpicture}
\end{flushleft}%
\vspace{-35pt}
\begin{flushleft}
\begin{tikzpicture}
    \definecolor{darkgray176}{RGB}{176,176,176}
    \definecolor{darkorange25512714}{RGB}{255,127,14}
    \definecolor{steelblue31119180}{RGB}{31,119,180}
    
    \begin{axis}[
    tick align=outside,
    tick pos=left,
    scaled y ticks=false,
    height=2.5cm,
    yticklabel style={text width=3em, align=right, /pgf/number format/fixed},
    width=0.9\linewidth,
    x grid style={darkgray176},
    xmin=10, xmax=20,
    xtick style={color=black},
    xtick=\empty,
    y grid style={darkgray176},
    ylabel={$\dot{p}_z$[m/s]},
    ymin=-0.01, ymax=0.13,
    ytick={0.00, 0.13},
    ]
\addplot [draw=none, fill=gray!50,opacity=0.5] coordinates {(13.8,-1) (13.8,5) (17,5) (17,-1)};
\addplot [semithick, steelblue31119180]
table {%
10.2398681640625 0.000854879374301378
10.3598666191101 0.00197260490879352
10.4798650741577 0.00308279087968128
10.5998635292053 0.00418513126638729
10.7198619842529 0.00527934979950743
10.8398604393005 0.00636519571819148
10.9598588943481 0.00744243959218218
11.0798573493958 0.00851086920737167
11.1998558044434 0.00957028551386705
11.319854259491 0.0106204986357311
11.4398527145386 0.0116613239417783
11.5598511695862 0.0126925781770545
11.6798496246338 0.0137140756549078
11.7998480796814 0.0147256245098667
11.919846534729 0.0157270230118741
12.0398449897766 0.0167180559427792
12.1598434448242 0.017698491036364
12.2798418998718 0.018668075483567
12.3998403549194 0.0196265325049702
12.519838809967 0.020573557993027
12.6398372650146 0.0215088172269245
12.7598357200623 0.0224319416634006
12.8798341751099 0.0233425258072585
12.9998326301575 0.0242401241657496
13.1198310852051 0.0251242482914136
13.2398295402527 0.0259943639183864
13.3598279953003 0.0268498881975917
13.4798264503479 0.027690187036634
13.5998249053955 0.0284989295880256
13.7198233604431 0.0292490635619868
13.8398218154907 0.0299397704051421
13.9598202705383 0.0305707256387052
14.0798187255859 0.0311416722428099
14.1998171806335 0.031652429162894
14.3198156356812 0.0321028993112531
14.4398140907288 0.0324930769651504
14.5598125457764 0.0328230544702932
14.679811000824 0.0330930281663112
14.7998094558716 0.0333033034590877
14.9198079109192 0.0334542989733569
15.0398063659668 0.0335465497278596
15.1598048210144 0.0335807092844865
15.279803276062 0.0335575508321827
15.3998017311096 0.0334779671758626
15.5198001861572 0.0333429696101333
15.6397986412048 0.0331536856671541
15.7597970962524 0.032911355737405
15.8797955513 0.0326173285714045
15.9997940063477 0.0322730556794333
16.1197924613953 0.0318800846549963
16.2397909164429 0.0314400514560236
16.3597893714905 0.0309546716855827
16.4797878265381 0.0304257309211005
16.5997862815857 0.0298550741476921
16.7197847366333 0.0292445943571299
16.8397831916809 0.0285962203792083
16.9597816467285 0.0279119040167325
17.0797801017761 0.0271936065590724
17.1997785568237 0.0264432847521546
17.3197770118713 0.0256628763049301
17.4397754669189 0.024854285013763
17.5597739219666 0.02401936558687
17.6797723770142 0.0231599082509469
17.7997708320618 0.0222778960403736
17.9197692871094 0.0213913610438115
18.039767742157 0.0205107483827932
18.1597661972046 0.0196364962519009
18.2797646522522 0.0187689837080836
18.3997631072998 0.0179085272860858
18.5197615623474 0.0170553777782807
18.639760017395 0.0162097171783986
18.7597584724426 0.015371655788386
18.8797569274902 0.0145412294874755
18.9997553825378 0.0137183971625063
19.1197538375854 0.0129030382986149
19.2397522926331 0.0120949507296102
19.3597507476807 0.0112938485476752
19.4797492027283 0.0104993601724989
19.5997476577759 0.00971102658053361
19.7197461128235 0.00892829969581356
19.8397445678711 0.0081505409446524
19.9597430229187 0.00737701997756964
20.0797414779663 0.00660691356298074
20.1997399330139 0.00583930465852657
20.3197383880615 0.0050731816674171
20.4397368431091 0.00430743788882378
20.5597352981567 0.00354087117317649
20.6797337532043 0.00277218379520585
20.799732208252 0.00199998255971915
20.9197306632996 0.00122277915740807
};
\addplot [semithick, darkorange25512714]
table {%
10.2398681640625 0.000713497653285409
10.3598666191101 0.00184256664518796
10.4798650741577 0.00296189253121993
10.5998635292053 0.00407129083624429
10.7198619842529 0.00517075660479227
10.8398604393005 0.00626037397041368
10.9598588943481 0.00734025377676229
11.0798573493958 0.0084104941006367
11.1998558044434 0.00947115815211287
11.319854259491 0.0105222644099634
11.4398527145386 0.0115637846430788
11.5598511695862 0.0125956464142983
11.6798496246338 0.0136177375924551
11.7998480796814 0.0146299112138812
11.919846534729 0.0156319896922037
12.0398449897766 0.0166237678675831
12.1598434448242 0.0176050147281924
12.2798418998718 0.0185754738533272
12.3998403549194 0.0195348627479843
12.519838809967 0.020482871290554
12.6398372650146 0.021419159521948
12.7598357200623 0.0223433549845429
12.8798341751099 0.023255049239555
12.9998326301575 0.024153570126402
13.1198310852051 0.0250371055785828
13.2398295402527 0.0259022933145968
13.3598279953003 0.0267443535891432
13.4798264503479 0.0275575358385657
13.5998249053955 0.028335683273758
13.7198233604431 0.0290727656477664
13.8398218154907 0.0297632943802763
13.9598202705383 0.0304150710107791
14.0798187255859 0.0279474982171224
14.1998171806335 0.0220287606126804
14.3198156356812 0.0155307101442041
14.4398140907288 0.00982301698213177
14.5598125457764 0.00540543856865113
14.679811000824 0.0023121724250052
14.7998094558716 0.000358723065816293
14.9198079109192 -0.000715019707310248
15.0398063659668 -0.0011677786511968
15.1598048210144 -0.00122061049826423
15.279803276062 -0.00104463269227759
15.3998017311096 -0.000761900445924466
15.5198001861572 -0.000452567843032127
15.6397986412048 -0.000164186116553314
15.7597970962524 7.91778600361817e-05
15.8797955513 0.000269128335443045
15.9997940063477 0.000406823442982083
16.1197924613953 0.000498527932018412
16.2397909164429 0.000552604489509839
16.3597893714905 0.000577639010713901
16.4797878265381 0.000581397581180485
16.5997862815857 0.000570353066920667
16.7197847366333 0.000549573220615418
16.8397831916809 0.000522816343393926
16.9597816467285 0.00271434506310295
17.0797801017761 0.0140870339036
17.1997785568237 0.0373162328752107
17.3197770118713 0.0708797894680803
17.4397754669189 0.100686575610713
17.5597739219666 0.119098891810739
17.6797723770142 0.125750621160199
17.7997708320618 0.12116768536369
17.9197692871094 0.107155067096411
18.039767742157 0.0891321119041311
18.1597661972046 0.0713540280376377
18.2797646522522 0.0558288741559017
18.3997631072998 0.0432146536488579
18.5197615623474 0.0334574086302166
18.639760017395 0.0261834114571801
18.7597584724426 0.0209161763235255
18.8797569274902 0.0171862429471447
18.9997553825378 0.0145809097819383
19.1197538375854 0.0127618996467871
19.2397522926331 0.0114663388840109
19.3597507476807 0.0104994005198365
19.4797492027283 0.00972330179437596
19.5997476577759 0.00904539749932618
19.7197461128235 0.00840696294683705
19.8397445678711 0.00777350842536782
19.9597430229187 0.00712695434759185
20.0797414779663 0.00645966089074342
20.1997399330139 0.00577011037189485
20.3197383880615 0.00505995051134096
20.4397368431091 0.00433208890095763
20.5597352981567 0.00358955447751214
20.6797337532043 0.00283488832021192
20.799732208252 0.00206987846626641
20.9197306632996 0.00129550264164988
};
\end{axis}

\end{tikzpicture}
\end{flushleft}%
\vspace{-35pt}
\begin{flushleft}
\begin{tikzpicture}

    \definecolor{darkgray176}{RGB}{176,176,176}
    \definecolor{darkorange25512714}{RGB}{255,127,14}
    \definecolor{steelblue31119180}{RGB}{31,119,180}
    
    \begin{axis}[
    tick align=outside,
    tick pos=left,
    scaled y ticks=false,
    height=2.5cm,
    yticklabel style={text width=3em, align=right, /pgf/number format/fixed},
    width=0.9\linewidth,
    x grid style={darkgray176},
    xmin=10, xmax=20,
    xtick style={color=black},
    xtick=\empty,
    y grid style={darkgray176},
    ylabel={$e_x$ [m]},
    ymin=0.2, ymax=0.7,
    ytick={0.2, 0.4, 0.6},
      trim axis left,
      trim axis right,
    ]
\addplot [draw=none, fill=gray!50,opacity=0.5] coordinates {(13.8,-1) (13.8,5) (17,5) (17,-1)};
\addplot [semithick, steelblue31119180]
table {%
10.2398681640625 0.499078086696579
10.3598666191101 0.499272890775358
10.4798650741577 0.499537433853841
10.5998635292053 0.499872545631504
10.7198619842529 0.500278535872349
10.8398604393005 0.500755204530514
10.9598588943481 0.501301850980159
11.0798573493958 0.501917282368475
11.1998558044434 0.502599821121979
11.319854259491 0.503347311647144
11.4398527145386 0.504157126276713
11.5598511695862 0.505026170523009
11.6798496246338 0.505950887709041
11.7998480796814 0.50692726305744
11.919846534729 0.507950827326115
12.0398449897766 0.509016660088086
12.1598434448242 0.510119392761251
12.2798418998718 0.511253211501868
12.3998403549194 0.512411860083219
12.519838809967 0.513588642888355
12.6398372650146 0.514776428152892
12.7598357200623 0.51596765160053
12.8798341751099 0.517154320620267
12.9998326301575 0.518328019140083
13.1198310852051 0.519479913357176
13.2398295402527 0.520600758489496
13.3598279953003 0.521680906717309
13.4798264503479 0.522710316486747
13.5998249053955 0.523678066343017
13.7198233604431 0.524570020208509
13.8398218154907 0.525371278322619
13.9598202705383 0.526067531953304
14.0798187255859 0.526645154390963
14.1998171806335 0.527091286971398
14.3198156356812 0.527393921858279
14.4398140907288 0.527541980893935
14.5598125457764 0.527525389818892
14.679811000824 0.527335147166699
14.7998094558716 0.526963387161111
14.9198079109192 0.526403435977359
15.0398063659668 0.525649860777212
15.1598048210144 0.524698510988222
15.279803276062 0.523546551369643
15.3998017311096 0.522192486489868
15.5198001861572 0.520636176331338
15.6397986412048 0.518878842836989
15.7597970962524 0.51692306731578
15.8797955513 0.51477277873168
15.9997940063477 0.5124332330087
16.1197924613953 0.509910983592288
16.2397909164429 0.507213843612445
16.3597893714905 0.504350840094519
16.4797878265381 0.501332160757805
16.5997862815857 0.498169094028121
16.7197847366333 0.494873962966832
16.8397831916809 0.491460053883961
16.9597816467285 0.487941540455796
17.0797801017761 0.484333404206864
17.1997785568237 0.480651352241454
17.3197770118713 0.476911733120571
17.4397754669189 0.47313145177604
17.5597739219666 0.469327884334337
17.6797723770142 0.46551879368892
17.7997708320618 0.461722242999452
17.9197692871094 0.457954655849348
18.039767742157 0.454227540495503
18.1597661972046 0.450551313765488
18.2797646522522 0.4469360565949
18.3997631072998 0.44339150367954
18.5197615623474 0.43992703498003
18.639760017395 0.436551668950599
18.7597584724426 0.433274057348511
18.8797569274902 0.430102481466485
18.9997553825378 0.427044849617455
19.1197538375854 0.424108695689096
19.2397522926331 0.421301178574712
19.3597507476807 0.418629082277256
19.4797492027283 0.416098816474415
19.5997476577759 0.413716417324795
19.7197461128235 0.411487548288221
19.8397445678711 0.409417500726959
19.9597430229187 0.407511194049313
20.0797414779663 0.405773175152386
20.1997399330139 0.404207616916912
20.3197383880615 0.402818315503889
20.4397368431091 0.401608686200336
20.5597352981567 0.400581757559807
20.6797337532043 0.399740163582491
20.799732208252 0.399086133679778
20.9197306632996 0.398621480169301
};
\addplot [semithick, darkorange25512714]
table {%
10.2398681640625 0.499155404333094
10.3598666191101 0.499345687257004
10.4798650741577 0.499605505555322
10.5998635292053 0.499935732968207
10.7198619842529 0.50033671906453
10.8398604393005 0.500808302387549
10.9598588943481 0.5013498214908
11.0798573493958 0.501960124115249
11.1998558044434 0.502637574832229
11.319854259491 0.503380061472268
11.4398527145386 0.504185000615511
11.5598511695862 0.505049342362077
11.6798496246338 0.505969574546833
11.7998480796814 0.506941726521332
11.919846534729 0.507961372598548
12.0398449897766 0.509023635242793
12.1598434448242 0.510123188085017
12.2798418998718 0.511254258849377
12.3998403549194 0.512410632287309
12.519838809967 0.513585653227964
12.6398372650146 0.51477222986678
12.7598357200623 0.515962837426105
12.8798341751099 0.51714952232951
12.9998326301575 0.518323898857249
13.1198310852051 0.519477073264172
13.2398295402527 0.520599494851131
13.3598279953003 0.521680831280576
13.4798264503479 0.522709924479462
13.5998249053955 0.523674846822565
13.7198233604431 0.524563048841773
13.8398218154907 0.525361573705022
13.9598202705383 0.526056644638985
14.0798187255859 0.526585059340443
14.1998171806335 0.526876328653324
14.3198156356812 0.526933717704406
14.4398140907288 0.52678627481171
14.5598125457764 0.526463272233869
14.679811000824 0.525987749406994
14.7998094558716 0.525376823955295
14.9198079109192 0.52464324904006
15.0398063659668 0.523796590411309
15.1598048210144 0.522843922705557
15.279803276062 0.521790285384392
15.3998017311096 0.520639074345154
15.5198001861572 0.519392427192465
15.6397986412048 0.518051591592586
15.7597970962524 0.516617248010483
15.8797955513 0.515089766476314
15.9997940063477 0.513469392124605
16.1197924613953 0.511756366230787
16.2397909164429 0.509950995715086
16.3597893714905 0.508053685503795
16.4797878265381 0.506064946634474
16.5997862815857 0.503985390238389
16.7197847366333 0.501815714619439
16.8397831916809 0.499556690127543
16.9597816467285 0.497324302065269
17.0797801017761 0.495585269113721
17.1997785568237 0.494480847635324
17.3197770118713 0.493742908921701
17.4397754669189 0.49272445764735
17.5597739219666 0.490646314341881
17.6797723770142 0.48696426901354
17.7997708320618 0.481574412996417
17.9197692871094 0.474884834222141
18.039767742157 0.467615341094183
18.1597661972046 0.460415785559642
18.2797646522522 0.453712586520035
18.3997631072998 0.447706666131259
18.5197615623474 0.442432106425017
18.639760017395 0.437824919758052
18.7597584724426 0.433778434390239
18.8797569274902 0.430179517383779
18.9997553825378 0.426928173341423
19.1197538375854 0.423945500207755
19.2397522926331 0.421174657881179
19.3597507476807 0.418578281679579
19.4797492027283 0.416134532748717
19.5997476577759 0.413833024906426
19.7197461128235 0.411671233435633
19.8397445678711 0.409651610368798
19.9597430229187 0.4077794226562
20.0797414779663 0.40606122912039
20.1997399330139 0.404503874248311
20.3197383880615 0.403113873136769
20.4397368431091 0.401897075092123
20.5597352981567 0.400858513454236
20.6797337532043 0.400002370502337
20.799732208252 0.399332005801409
20.9197306632996 0.39885001268608
};
\end{axis}

\end{tikzpicture}
\end{flushleft}%
\vspace{-35pt}
\begin{flushleft}
\begin{tikzpicture}

    \definecolor{darkgray176}{RGB}{176,176,176}
    \definecolor{darkorange25512714}{RGB}{255,127,14}
    \definecolor{steelblue31119180}{RGB}{31,119,180}
    
    \begin{axis}[
    tick align=outside,
    tick pos=left,
    scaled y ticks=false,
    height=2.5cm,
    yticklabel style={text width=2.98em, align=right, /pgf/number format/fixed},
    width=0.9\linewidth,
    x grid style={darkgray176},
    xlabel=\small{Time [s]},
    xmin=10, xmax=20,
    xtick style={color=black},
    y grid style={darkgray176},
    ylabel={$e_y$ [m]},
    ymin=-0.25, ymax=0.25,
    ytick style={color=black},
    ytick={-0.2, 0, 0.2},
      trim axis left,
      trim axis right,
    legend style={legend cell align={left},font=\tiny,at={(0.0,-0.7)},anchor=north},,
    legend columns=-1
    ]
\addplot [draw=none, fill=gray!50,opacity=0.5,forget plot] coordinates {(13.8,-1) (13.8,5) (17,5) (17,-1)};
\addplot [semithick, steelblue31119180]
table {%
10.2398681640625 -0.201779300477391
10.3598666191101 -0.201516156233448
10.4798650741577 -0.201020239894063
10.5998635292053 -0.200293341673541
10.7198619842529 -0.199337092069728
10.8398604393005 -0.198152960313136
10.9598588943481 -0.196742254833186
11.0798573493958 -0.195106125674911
11.1998558044434 -0.193245568806299
11.319854259491 -0.191161432262954
11.4398527145386 -0.188854424082672
11.5598511695862 -0.18632512198789
11.6798496246338 -0.183573984778702
11.7998480796814 -0.180601365403116
11.919846534729 -0.177407525674473
12.0398449897766 -0.173992652608333
12.1598434448242 -0.170356876352677
12.2798418998718 -0.166500289685837
12.3998403549194 -0.162422969056236
12.519838809967 -0.158124997136574
12.6398372650146 -0.153606486862727
12.7598357200623 -0.148867606924049
12.8798341751099 -0.143908608667208
12.9998326301575 -0.13872985436992
13.1198310852051 -0.133331846834077
13.2398295402527 -0.127715260239755
13.3598279953003 -0.121880972192427
13.4798264503479 -0.115830096885431
13.5998249053955 -0.109564956151947
13.7198233604431 -0.103094033103282
13.8398218154907 -0.0964285887881112
13.9598202705383 -0.0895805069823779
14.0798187255859 -0.0825622885356279
14.1998171806335 -0.0753870403594479
14.3198156356812 -0.068068454911312
14.4398140907288 -0.0606207804320031
14.5598125457764 -0.0530587822968624
14.679811000824 -0.0453976959418727
14.7998094558716 -0.037653171923571
14.9198079109192 -0.029841213765456
15.0398063659668 -0.021978109331418
15.1598048210144 -0.0140803565472744
15.279803276062 -0.00616458436332986
15.3998017311096 0.00175253008727188
15.5198001861572 0.00965434712895807
15.6397986412048 0.0175243519388572
15.7597970962524 0.0253462426268923
15.8797955513 0.0331040179187241
15.9997940063477 0.0407820633607394
16.1197924613953 0.048365234981242
16.2397909164429 0.0558389393703836
16.3597893714905 0.0631892091826643
16.4797878265381 0.0704027731192676
16.5997862815857 0.0774671195121078
16.7197847366333 0.0843705527060239
16.8397831916809 0.0911022415187087
16.9597816467285 0.097652259148133
17.0797801017761 0.104011613992799
17.1997785568237 0.110172270949361
17.3197770118713 0.116127162853241
17.4397754669189 0.121870191829054
17.5597739219666 0.12739622041722
17.6797723770142 0.13270105243943
17.7997708320618 0.137781406146472
17.9197692871094 0.142636102426809
18.039767742157 0.147267790725039
18.1597661972046 0.151679904750507
18.2797646522522 0.155876073077681
18.3997631072998 0.15986008175899
18.5197615623474 0.163635836691053
18.639760017395 0.167207325884809
18.7597584724426 0.17057858177947
18.8797569274902 0.17375364372948
18.9997553825378 0.176736520782908
19.1197538375854 0.179531154859222
19.2397522926331 0.182141384424286
19.3597507476807 0.184570908751014
19.4797492027283 0.186823252845468
19.5997476577759 0.188901733110646
19.7197461128235 0.190809423813788
19.8397445678711 0.192549124418122
19.9597430229187 0.194123327836541
20.0797414779663 0.195534189663151
20.1997399330139 0.196783498438925
20.3197383880615 0.197872647010192
20.4397368431091 0.198802605043391
20.5597352981567 0.199573892766753
20.6797337532043 0.200186556019327
20.799732208252 0.200640142700346
20.9197306632996 0.200933680727382
};
\addlegendentry{Original}
\addplot [semithick, darkorange25512714]
table {%
10.2398681640625 -0.201585166672581
10.3598666191101 -0.201351461122897
10.4798650741577 -0.200882577982261
10.5998635292053 -0.20018081789743
10.7198619842529 -0.199248273702555
10.8398604393005 -0.198086806087202
10.9598588943481 -0.196698037217484
11.0798573493958 -0.195083356513364
11.1998558044434 -0.193243934053461
11.319854259491 -0.191180738307899
11.4398527145386 -0.188894555972806
11.5598511695862 -0.186386012544989
11.6798496246338 -0.183655592924637
11.7998480796814 -0.180703661787025
11.919846534729 -0.177530483754088
12.0398449897766 -0.174136243560325
12.1598434448242 -0.170521066480085
12.2798418998718 -0.166685039295024
12.3998403549194 -0.162628232055353
12.519838809967 -0.158350720844174
12.6398372650146 -0.15385261170281
12.7598357200623 -0.149134065824212
12.8798341751099 -0.144195326079891
12.9998326301575 -0.139036757257239
13.1198310852051 -0.133659005688974
13.2398295402527 -0.128063314267642
13.3598279953003 -0.122251876977204
13.4798264503479 -0.116228148908627
13.5998249053955 -0.109997062455943
13.7198233604431 -0.103565133707057
13.8398218154907 -0.0969404669233932
13.9598202705383 -0.0901325913837425
14.0798187255859 -0.0831522748752718
14.1998171806335 -0.076027327309002
14.3198156356812 -0.0687994133095549
14.4398140907288 -0.0615086646228856
14.5598125457764 -0.0541874537219911
14.679811000824 -0.0468597215182393
14.7998094558716 -0.039542223770525
14.9198079109192 -0.0322461553341077
15.0398063659668 -0.0249786422504524
15.1598048210144 -0.0177439797942237
15.279803276062 -0.0105446091323524
15.3998017311096 -0.00338185332920261
15.5198001861572 0.00374355736604844
15.6397986412048 0.0108311352035292
15.7597970962524 0.0178804211097007
15.8797955513 0.0248908702017909
15.9997940063477 0.0318618042458537
16.1197924613953 0.0387924052847964
16.2397909164429 0.0456817315759953
16.3597893714905 0.0525287431264829
16.4797878265381 0.0593323290040415
16.5997862815857 0.0660913321600652
16.7197847366333 0.0728045698851869
16.8397831916809 0.0794708494841281
16.9597816467285 0.0862167095655957
17.0797801017761 0.0934633475888534
17.1997785568237 0.101297886757301
17.3197770118713 0.109520774739698
17.4397754669189 0.11777587114496
17.5597739219666 0.125653510573245
17.6797723770142 0.13280845170667
17.7997708320618 0.139046042461197
17.9197692871094 0.144373963600815
18.039767742157 0.148964254358043
18.1597661972046 0.153032211672457
18.2797646522522 0.156762953738282
18.3997631072998 0.160284848804481
18.5197615623474 0.16367061802545
18.639760017395 0.166950671829935
18.7597584724426 0.170128565258432
18.8797569274902 0.173193807299771
18.9997553825378 0.176130718196632
19.1197538375854 0.178923711190094
19.2397522926331 0.181559929673433
19.3597507476807 0.184030168898636
19.4797492027283 0.186328808794564
19.5997476577759 0.188453256698048
19.7197461128235 0.190403211920034
19.8397445678711 0.192179930517723
19.9597430229187 0.19378558055622
20.0797414779663 0.195222723333192
20.1997399330139 0.196493923928735
20.3197383880615 0.197601477285525
20.4397368431091 0.198547228479352
20.5597352981567 0.199332464251928
20.6797337532043 0.199957854723731
20.799732208252 0.200423427749103
20.9197306632996 0.200728562462671
};
\addlegendentry{Modified}
\end{axis}

\end{tikzpicture}
\end{flushleft}%
\vspace{-30pt}
  \end{center}
\caption{Demonstration of trajectory modification from human feedback. The robot was programmed to move diagonally from the bottom left to the top right. In the first plot, it can be seen that a contact occurred between 13.8s and 17s. During the contact, the skin-enabled controller accurately localized the contact point at the top of the robot and reduced the velocity of the contact point to 0 in the z-direction (shown as $p_z$ in the second plot). The original trajectory of the robot at the end effector was maintained in the x and y directions, namely $e_x$ and $e_y$, as shown in the third and fourth plots respectively.}
  \label{fig: ssa demo}
  \end{center}
\end{figure}
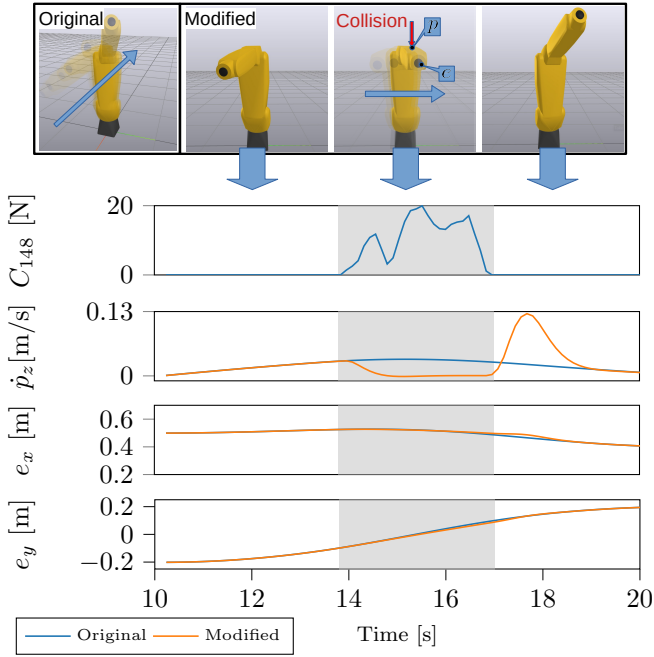
\setlength{\textfloatsep}{10pt}

To demonstrate the effectiveness of \eqref{eq: ccta}, we conducted the following experiment when setting $f(F_{i})\equiv 0$. The robot's reference trajectory is to move its end effector from the bottom left to the top right, as shown in Figure~\ref{fig: ssa demo}. When human touches the robot, the skin successfully localized the point of contact to the 184th sensor cell, and the robot's trajectory was immediately modified to reduce the Cartesian velocity in the direction of the contact to zero. 
Extensive evaluation of the proposed algorithm in \eqref{eq: ccta} will be left for future work. 

\subsection{Skin-Enabled Admittance Control}
This use case discusses how to let the robot generate desired force feedback to the human during contact, which is an important capability during human-robot interactions~\citep{shek2022learning}.  
However, prior works only utilized joint torque sensors, which can not localize the contact location, hence unable to correctly regulate the contact force. 
Skin-enabled admittance control solves this limitation by dynamically regulate the forces applied to the contact point (\cite{touchadmittance}). 
The corresponding dynamic equation is:
\begin{equation}
  M\Delta\ddot{x}(t) + B\Delta\dot{x}(t) + K\Delta{x(t)} = F_{i}(t) \label{second order system}
\end{equation}
where $\Delta{x}(t)$ is the contact point displacement, $\Delta{\dot{x}}(t)$ and $ \Delta{\ddot{x}}(t)$ are its first and second derivatives, $F_{i}(t)$ denotes the measured contact force from the skin. The parameters $M$ (inertia), $B$ (damping coefficient), and $K$ (stiffness) are design parameters to be selected case-by-case. 



\begin{figure}[t]
  \begin{center}
  \centering
  \begin{center}
  \centering
  \includegraphics[width=0.9\linewidth]{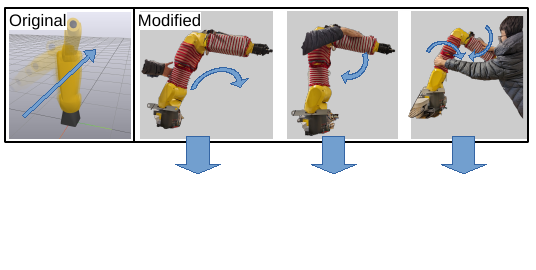}
\vspace{-55pt}
\begin{flushleft}
\begin{tikzpicture}

    \definecolor{darkgray176}{RGB}{176,176,176}
    \definecolor{steelblue31119180}{RGB}{31,119,180}
    
    \begin{axis}[
    tick align=outside,
    tick pos=left,
    scaled y ticks=false,
    height=2.5cm,
    yticklabel style={text width=2.5em, align=right, /pgf/number format/fixed},
    width=0.9\linewidth,
    x grid style={darkgray176},
    xmin=4, xmax=19.9,
    xtick=\empty,
    y grid style={darkgray176},
    ylabel={$C_{13}$ [N]},
    ytick={0,20},
    ymin=0, ymax=20,
    ]
\addplot [draw=none, fill=gray!50,opacity=0.5] coordinates {(6.8,-1) (6.8,25) (9.6,25) (9.6,-1)};
\addplot [draw=none, fill=gray!50,opacity=0.5] coordinates {(11.8,-1) (11.8,25) (13.6,25) (13.6,-1)};
\addplot [draw=none, fill=gray!50,opacity=0.5] coordinates {(16.1,-1) (16.1,25) (19.8,25) (19.8,-1)};
\addplot [semithick, steelblue31119180]
table {%
3.99994850158691 0
4.11994695663452 0
4.23994541168213 0
4.35994386672974 0
4.47994232177734 0
4.59994077682495 0
4.71993923187256 0
4.83993768692017 0
4.95993614196777 0
5.07993459701538 0
5.19993305206299 0
5.3199315071106 0
5.4399299621582 0
5.55992841720581 0
5.67992687225342 0
5.79992532730103 0
5.91992378234863 0
6.03992223739624 0
6.15992069244385 0
6.27991914749146 0
6.39991760253906 0
6.51991605758667 0
6.63991451263428 0
6.75991296768188 0
6.87991142272949 0.11083936268688
6.9999098777771 8.74674285384009
7.11990833282471 9.68991923699989
7.23990678787231 12.2034538378585
7.35990524291992 12.636567167501
7.47990369796753 12.427552261257
7.59990215301514 13.1435193888454
7.71990060806274 13.4708510655172
7.83989906311035 13.8555691273837
7.95989751815796 13.1455175956502
8.07989597320557 14.0517081287088
8.19989442825317 14.1533953423188
8.31989288330078 14.6897172884577
8.43989133834839 15.1032835952188
8.559889793396 15.2747485472812
8.6798882484436 15.5544307901064
8.79988670349121 15.2168679579895
8.91988515853882 14.3907277736051
9.03988361358643 14.1638604548994
9.15988206863403 14.6149430035125
9.27988052368164 15.0457095684644
9.39987897872925 4.47510831878565
9.51987743377686 0
9.63987588882446 0
9.75987434387207 0
9.87987279891968 0
9.99987125396729 0
10.1198697090149 0
10.2398681640625 0
10.3598666191101 0
10.4798650741577 0
10.5998635292053 0
10.7198619842529 0
10.8398604393005 0
10.9598588943481 0
11.0798573493958 0
11.1998558044434 0
11.319854259491 0
11.4398527145386 0
11.5598511695862 0
11.6798496246338 0
11.7998480796814 0
11.919846534729 0
12.0398449897766 0
12.1598434448242 0
12.2798418998718 0
12.3998403549194 0
12.519838809967 0
12.6398372650146 0
12.7598357200623 0
12.8798341751099 0
12.9998326301575 0
13.1198310852051 0
13.2398295402527 0
13.3598279953003 0
13.4798264503479 0
13.5998249053955 0
13.7198233604431 0
13.8398218154907 0
13.9598202705383 0
14.0798187255859 0
14.1998171806335 0
14.3198156356812 0
14.4398140907288 0
14.5598125457764 0
14.679811000824 0
14.7998094558716 0
14.9198079109192 0
15.0398063659668 0
15.1598048210144 0
15.279803276062 0
15.3998017311096 0
15.5198001861572 0
15.6397986412048 0
15.7597970962524 0
15.8797955513 0
15.9997940063477 0
16.1197924613953 0
16.2397909164429 0
16.3597893714905 0
16.4797878265381 0.107095028923159
16.5997862815857 0.05364106689375
16.7197847366333 7.33903491935025
16.8397831916809 12.07165734057
16.9597816467285 12.7914045745007
17.0797801017761 12.5705606195586
17.1997785568237 11.3483233254115
17.3197770118713 13.4335113740299
17.4397754669189 13.4581001546699
17.5597739219666 12.3792586975212
17.6797723770142 13.5563207703484
17.7997708320618 13.1491950408989
17.9197692871094 13.9714377222007
18.039767742157 14.5495381682414
18.1597661972046 15.5254836370312
18.2797646522522 14.4497224776763
18.3997631072998 15.2187926949985
18.5197615623474 15.043421223715
18.639760017395 14.871495724642
18.7597584724426 15.597075323348
18.8797569274902 14.4309657985746
18.9997553825378 15.7461564190764
19.1197538375854 15.7907611950946
19.2397522926331 15.0309117843592
19.3597507476807 15.6511954276997
19.4797492027283 15.4814971304
19.5997476577759 15.1215159829414
19.7197461128235 0.56624991929525
19.8397445678711 0
19.9597430229187 0
};
\end{axis}
\end{tikzpicture}
\end{flushleft}%
\vspace{-35pt}
\begin{flushleft}
\begin{tikzpicture}

    \definecolor{darkgray176}{RGB}{176,176,176}
    \definecolor{steelblue31119180}{RGB}{31,119,180}
    
    \begin{axis}[
    tick align=outside,
    tick pos=left,
    scaled y ticks=false,
    height=2.5cm,
    yticklabel style={text width=2.5em, align=right, /pgf/number format/fixed},
    width=0.9\linewidth,
    x grid style={darkgray176},
    xmin=4, xmax=19.9,
    xtick=\empty,
    y grid style={darkgray176},
    ylabel={$C_{153}$ [N]},
    ytick={0,20},
    ymin=0, ymax=20,
    ]
\addplot [draw=none, fill=gray!50,opacity=0.5] coordinates {(6.8,-1) (6.8,25) (9.6,25) (9.6,-1)};
\addplot [draw=none, fill=gray!50,opacity=0.5] coordinates {(11.8,-1) (11.8,25) (13.6,25) (13.6,-1)};
\addplot [draw=none, fill=gray!50,opacity=0.5] coordinates {(16.1,-1) (16.1,25) (19.8,25) (19.8,-1)};
\addplot [semithick, steelblue31119180]
table {%
3.99994850158691 0
4.11994695663452 0
4.23994541168213 0
4.35994386672974 0
4.47994232177734 0
4.59994077682495 0
4.71993923187256 0
4.83993768692017 0
4.95993614196777 0
5.07993459701538 0
5.19993305206299 0
5.3199315071106 0
5.4399299621582 0
5.55992841720581 0
5.67992687225342 0
5.79992532730103 0
5.91992378234863 0
6.03992223739624 0
6.15992069244385 0
6.27991914749146 0
6.39991760253906 0
6.51991605758667 0
6.63991451263428 0
6.75991296768188 0
6.87991142272949 0
6.9999098777771 0
7.11990833282471 0
7.23990678787231 0
7.35990524291992 0
7.47990369796753 0
7.59990215301514 0
7.71990060806274 0.0178225260789318
7.83989906311035 0
7.95989751815796 0
8.07989597320557 0
8.19989442825317 0
8.31989288330078 0
8.43989133834839 0
8.559889793396 0
8.6798882484436 0
8.79988670349121 0
8.91988515853882 0
9.03988361358643 0
9.15988206863403 0
9.27988052368164 0
9.39987897872925 0
9.51987743377686 0
9.63987588882446 0
9.75987434387207 0
9.87987279891968 0
9.99987125396729 0
10.1198697090149 0
10.2398681640625 0
10.3598666191101 0
10.4798650741577 0
10.5998635292053 0
10.7198619842529 0
10.8398604393005 0
10.9598588943481 0
11.0798573493958 0
11.1998558044434 0
11.319854259491 0
11.4398527145386 0
11.5598511695862 0
11.6798496246338 0.0184578816386509
11.7998480796814 0
11.919846534729 0.0322404695649152
12.0398449897766 6.40560524567735
12.1598434448242 15.0954446883503
12.2798418998718 15.1874589399116
12.3998403549194 15.6148025738872
12.519838809967 15.7304318979674
12.6398372650146 17.6819105607995
12.7598357200623 18.0092192662137
12.8798341751099 18.0299065455497
12.9998326301575 19.4002936067686
13.1198310852051 19.3418351636944
13.2398295402527 20.3907303667329
13.3598279953003 14.2392383067139
13.4798264503479 0.0133513965146813
13.5998249053955 0
13.7198233604431 0
13.8398218154907 0
13.9598202705383 0
14.0798187255859 0
14.1998171806335 0
14.3198156356812 0
14.4398140907288 0
14.5598125457764 0
14.679811000824 0
14.7998094558716 0
14.9198079109192 0
15.0398063659668 0
15.1598048210144 0
15.279803276062 0
15.3998017311096 0
15.5198001861572 0
15.6397986412048 0.0206093906467186
15.7597970962524 0.026140188668285
15.8797955513 0
15.9997940063477 0.0305146474622962
16.1197924613953 0.114834171587167
16.2397909164429 2.00973153281177
16.3597893714905 9.98244892673317
16.4797878265381 8.71280984712224
16.5997862815857 0.22454284384934
16.7197847366333 7.78386812466265
16.8397831916809 13.1242444814563
16.9597816467285 4.78935236382691
17.0797801017761 14.4817472113797
17.1997785568237 15.3719997196635
17.3197770118713 15.6067916580851
17.4397754669189 16.1034944124775
17.5597739219666 16.9828087253454
17.6797723770142 16.7485948919052
17.7997708320618 17.3654488007074
17.9197692871094 16.5688716141171
18.039767742157 16.9191151948815
18.1597661972046 17.0738549000115
18.2797646522522 17.1612130255546
18.3997631072998 17.3042681197048
18.5197615623474 16.2214303086435
18.639760017395 15.1327168762105
18.7597584724426 15.2105691246434
18.8797569274902 14.4243852602543
18.9997553825378 16.8006026160468
19.1197538375854 8.62527045367005
19.2397522926331 0
19.3597507476807 0
19.4797492027283 0
19.5997476577759 0
19.7197461128235 0
19.8397445678711 0
19.9597430229187 0
};
\end{axis}
\end{tikzpicture}
\end{flushleft}%
\vspace{-35pt}
\include{admit_j2}%
\vspace{-35pt}
\include{admit_j3}%
\vspace{-20pt}
  \end{center}
\caption{Demonstration of robot admittance control with the skin. The robot moves from the bottom left to the top right, while the human operator intervenes at different stages. The lower portion of the robot is pushed from the back at first, followed by pressing the top link downwards, and finally simultaneous pushing and pressing. The three intervention stages are highlighted in gray. This scenario involves multiple joints and contacts between the human and robot. The results indicate that interactions with the top link (triggering $C_{153}$ and moving $q_3$) and the lower link (triggering $C_{13}$ and moving $q_2$) can be performed independently of each other.}
  \label{fig: admittance demo}
  \end{center}
\end{figure}
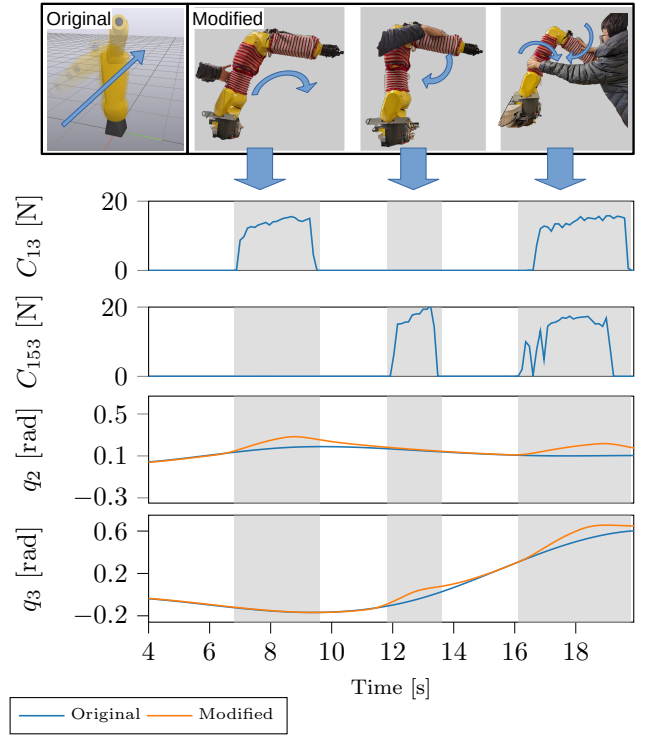
\setlength{\textfloatsep}{10pt}

As the robot system operates on a discrete domain, admittance control law in \eqref{second order system} is converted from time domain to Z domain using Laplace transform and Tustin's approximation~\citep{lahr2016understanding}. 
The admittance control displacements at time step $k$ is: 
\begin{equation} 
\begin{array}{ll}
\Delta{x}(k) &= [T_s^2F_{i}(k) + 2T_s^2F_{i}(k-1)\\
            & + T_s^2F_{i}(k-2) - (2KT_s^2 - 8M)\Delta{x}(k-1)\\
            &- (4M-2BT_s^2+KT_s^2)\Delta{x}(k-2)]\\
            &* 1/[4M+2BT_s^2+KT_s^2]
\end{array}
\label{admittance control equation}
\end{equation}
where $T_s$ is the sample time of the control loop.

The algorithm \eqref{admittance control equation} is tested on the FANUC LR Mate 200id/7L industrial robot with $K=500 N/rad$, a critically damped damping parameter of $B=2\sqrt{K}$, and $M=0$. Unlike joint torque sensors, which cannot distinguish forces applied to different joints in the multi-joint, multi-contact case, the skin-enabled admittance control measure contact force and location on the robot surface and allows each joint to respond independently to contacts. Figure \ref{fig: admittance demo} illustrates this capability. With skin-enabled admittance control, the interactive force profiles between humans and robots are smoother (compared to Fig. \ref{fig: ssa demo}). The human operator can interact with the robot in a more intuitive manner and provide rich contact information for learning from human feedback. 

It is worth noting that since admittance control works by adding the displacement of the robot state to the reference trajectory, and the tactile skin could sense multi-touch, a joint could respond to multiple contacts on a single joint. The desired robot trajectory in multi-contact scenarios can be calculated by equation~(\ref{multi desired trajetcory}). Testing of the algorithm will be left for future work.
\begin{equation} \label{multi desired trajetcory}
\begin{array}{ll}
x_{des}(k) = &x_{ref}(k) + \Delta{x}_{c1}(k) + \\
    &\Delta{x}_{c2}(k) + ... +\Delta{x}_{cN}(k).
\end{array}
\end{equation}


\section{Conclusion}

This work presents a comprehensive solution for rapidly fabricating, calibrating, and deploying textile and tactile skins for interactive industrial robots. We developed an automated skin calibration method that enables simultaneous calibration of contact force and locations for different robots, making it more efficient to deploy these skins in various industrial settings. Furthermore, we demonstrated the effectiveness of textile and tactile skins in control applications, which can help improve human-robot interaction and learning in real-world scenarios.

Our experimental results showed that calibrated textile and tactile skins can effectively sense contact locations and forces, demonstrating the potential of these skins as a valuable sensing modality for enhancing robot learning and control. For future work, we plan to streamline the calibration process for multiple skins and develop an integrated skin for multiple robot links. We also aim to utilize the vast information collected from the skin during human-robot interactions for more complex robot learning applications. 

\begin{ack}
This research is supported by the CMU Manufacturing Futures Institute, made possible by the Richard King Mellon Foundation.
\end{ack}

\bibliography{ifacconf}                              

\begin{thebibliography}{16}
\providecommand{\natexlab}[1]{#1}
\providecommand{\url}[1]{\texttt{#1}}
\providecommand{\urlprefix}{URL }
\expandafter\ifx\csname urlstyle\endcsname\relax
  \providecommand{\doi}[1]{doi:\discretionary{}{}{}#1}\else
  \providecommand{\doi}{doi:\discretionary{}{}{}\begingroup
  \urlstyle{rm}\Url}\fi

\bibitem[{Baykas et~al.(2020)Baykas, Bayraktar, and Yigit}]{baykas2020safe}
Baykas, P.B., Bayraktar, E., and Yigit, C.B. (2020).
\newblock Safe human-robot interaction using variable stiffness,
  hyper-redundancy, and smart robotic skins.
\newblock In \emph{Service Robotics}. IntechOpen.

\bibitem[{Cao et~al.(2021)Cao, Laws, and y~Baena}]{cao2021six}
Cao, M.Y., Laws, S., and y~Baena, F.R. (2021).
\newblock Six-axis force/torque sensors for robotics applications: A review.
\newblock \emph{IEEE Sensors Journal}, 21(24), 27238--27251.

\bibitem[{Cirillo et~al.(2015)Cirillo, Ficuciello, Natale, Pirozzi, and
  Villani}]{cirillo2015conformable}
Cirillo, A., Ficuciello, F., Natale, C., Pirozzi, S., and Villani, L. (2015).
\newblock A conformable force/tactile skin for physical human--robot
  interaction.
\newblock \emph{IEEE Robotics and Automation Letters}, 1(1), 41--48.

\bibitem[{Dahiya et~al.(2013)Dahiya, Mittendorfer, Valle, Cheng, and
  Lumelsky}]{dahiya2013directions}
Dahiya, R.S., Mittendorfer, P., Valle, M., Cheng, G., and Lumelsky, V.J.
  (2013).
\newblock Directions toward effective utilization of tactile skin: A review.
\newblock \emph{IEEE Sensors Journal}, 13(11), 4121--4138.

\bibitem[{Fan et~al.(2022)Fan, Lee, Jackel, Howard, Lee, and
  Isler}]{fan2022enabling}
Fan, X., Lee, D., Jackel, L., Howard, R., Lee, D., and Isler, V. (2022).
\newblock Enabling low-cost full surface tactile skin for human robot
  interaction.
\newblock \emph{IEEE Robotics and Automation Letters}, 7(2), 1800--1807.

\bibitem[{Kloss et~al.(2020)Kloss, Bauza, Wu, Tenenbaum, Rodriguez, and
  Bohg}]{kloss2020accurate}
Kloss, A., Bauza, M., Wu, J., Tenenbaum, J.B., Rodriguez, A., and Bohg, J.
  (2020).
\newblock Accurate vision-based manipulation through contact reasoning.
\newblock In \emph{2020 IEEE International Conference on Robotics and
  Automation (ICRA)}, 6738--6744. IEEE.

\bibitem[{Lahr et~al.(2016)Lahr, Soares, Garcia, Siqueira, and
  Caurin}]{lahr2016understanding}
Lahr, G.J., Soares, J.V., Garcia, H.B., Siqueira, A.A., and Caurin, G.A.
  (2016).
\newblock Understanding the implementation of impedance control in industrial
  robots.
\newblock In \emph{2016 XIII Latin American Robotics Symposium and IV Brazilian
  Robotics Symposium (LARS/SBR)}, 269--274. IEEE.

\bibitem[{Lin et~al.(2017)Lin, Liu, Fan, and Tomizuka}]{liu2017ccta}
Lin, H.C., Liu, C., Fan, Y., and Tomizuka, M. (2017).
\newblock Real-time collision avoidance algorithm on industrial manipulators.
\newblock In \emph{2017 IEEE Conference on Control Technology and Applications
  (CCTA)}, 1294--1299.
\newblock \doi{10.1109/CCTA.2017.8062637}.

\bibitem[{Mohammadi et~al.(2019)Mohammadi, Xu, Tan, Choong, and
  Oetomo}]{moham2019sensors}
Mohammadi, A., Xu, Y., Tan, Y., Choong, P., and Oetomo, D. (2019).
\newblock Magnetic-based soft tactile sensors with deformable continuous force
  transfer medium for resolving contact locations in robotic grasping and
  manipulation.
\newblock \emph{Sensors}, 19(22).
\newblock \doi{10.3390/s19224925}.
\newblock \urlprefix\url{https://www.mdpi.com/1424-8220/19/22/4925}.

\bibitem[{Perron and Furnon()}]{ortools}
Perron, L. and Furnon, V. (????).
\newblock Or-tools.
\newblock \urlprefix\url{https://developers.google.com/optimization/}.

\bibitem[{Pugach et~al.(2016)Pugach, Melnyk, Tolochko, Pitti, and
  Gaussier}]{touchadmittance}
Pugach, G., Melnyk, A., Tolochko, O., Pitti, A., and Gaussier, P. (2016).
\newblock Touch-based admittance control of a robotic arm using neural learning
  of an artificial skin.
\newblock In \emph{2016 IEEE/RSJ International Conference on Intelligent Robots
  and Systems (IROS)}, 3374--3380.
\newblock \doi{10.1109/IROS.2016.7759519}.

\bibitem[{Shek et~al.(2023)Shek, Chen, and Liu}]{shek2022learning}
Shek, A., Chen, R., and Liu, C. (2023).
\newblock Learning from physical human feedback: An object-centric one-shot
  adaptation method.
\newblock \emph{arXiv preprint arXiv:2203.04951}.

\bibitem[{Si et~al.(2023)Si, Yu, Morozov, McCann, and
  Yuan}]{si2023robotsweater}
Si, Z., Yu, T.C., Morozov, K., McCann, J., and Yuan, W. (2023).
\newblock Robotsweater: Scalable, generalizable, and customizable
  machine-knitted tactile skins for robots.
\newblock \emph{arXiv preprint arXiv:2303.02858}.

\bibitem[{{Touch Solution}(2023)}]{examplewebsite}
{Touch Solution} (2023).
\newblock Touch solution.
\newblock \url{https://www.touche.solutions}.
\newblock Accessed: April 10, 2023.

\bibitem[{Tsuji and Kohama(2020)}]{tsuji2020general}
Tsuji, S. and Kohama, T. (2020).
\newblock A general-purpose safety light curtain using tof sensor for end
  effector on human collaborative robot.
\newblock \emph{IEEJ Transactions on Electrical and Electronic Engineering},
  15(12), 1868--1874.

\bibitem[{Wahrburg et~al.(2017)Wahrburg, B{\"o}s, Listmann, Dai, Matthias, and
  Ding}]{wahrburg2017motor}
Wahrburg, A., B{\"o}s, J., Listmann, K.D., Dai, F., Matthias, B., and Ding, H.
  (2017).
\newblock Motor-current-based estimation of cartesian contact forces and
  torques for robotic manipulators and its application to force control.
\newblock \emph{IEEE Transactions on Automation Science and Engineering},
  15(2), 879--886.

\end{thebibliography}
\appendix
\end{document}